\documentclass[journal,twoside,web]{ieeecolor}
\usepackage{tmi}
\usepackage{cite}
\usepackage{amsmath,amssymb,amsfonts}
\usepackage{algorithmic}
\usepackage{graphicx}
\usepackage{textcomp}
\usepackage{booktabs}
\usepackage{algorithm}
\usepackage{bbding}

\def\BibTeX{{\rm B\kern-.05em{\sc i\kern-.025em b}\kern-.08em
    T\kern-.1667em\lower.7ex\hbox{E}\kern-.125emX}}
\begin{document}
\title{Multiplex-Detection Based Multiple Instance Learning Network for Whole Slide Image Classification}
\author{Zhikang Wang, Yue Bi, Tong Pan, Xiaoyu Wang, Chris Bain, Richard Bassed, Seiya Imoto, Jianhua Yao, Jiangning Song
\thanks{This manuscript is submitted on 1 August 2022.}
\thanks{Zhikang Wang, Yue Bi, Tong Pan, Xiaoyu Wang and Jiangning Song (A/Prof) are with the Monash Biomedicine Discovery Institute and the Department of Biochemistry and Molecular Biology, Monash University. (e-mail: \{Zhikang.Wang; Yue.Bi; Tong.Pan; Tong.Pan2, Jiangning.Song\}@monash.edu)}
\thanks{Chris Bain (Prof) is with the Faculty of Information Technology, Monash University. (e-mail: Chris.A.Bain@monash.edu)}
\thanks{Richard Bassed (A/Prof) is with the Victorian Institute of Forensic Medicine, Australia. (e-mail: Richard.Bassed@vifm.org)}
\thanks{Seiya Imoto (Prof) is with the Human Genome Center, Institute of Medical Science, University of Tokyo, Japan. (e-mail:imoto@ims.u-tokyo.ac.jp)}
\thanks{Jianhua Yao, (Prof) is with the Tencent AI Lab, Shenzhen, China. (e-mail: jianhuayao@tencent.com)}
}

\maketitle

\begin{abstract}
Multiple instance learning (MIL) is a powerful technique to classify whole slide images (WSIs) for diagnostic pathology. 
The key challenge of MIL on WSI classification is to discover the \textit{critical instances} that trigger the bag label from the enormous amounts of samples. 
Previous methods are primarily designed under the independent and identical distribution hypothesis (\textit{i.i.d}), ignoring either the correlations between instances or heterogeneity of tumor tissues. 
In this paper, we propose a novel multiplex-detection-based multiple instance learning (MDMIL) to tackle the above issues. 
Specifically,  MDMIL is constructed by the internal query generation module (IQGM) and the multiplex detection module (MDM) and assisted by the memory-based contrastive loss in the training phase. 
Firstly, IQGM gives the rough probability of instances and generates the internal query (\textit{IQ}) for the subsequent MDM by analyzing and aggregating highly reliable features. Secondly, the multiplex-detection cross-attention (MDCA) and multi-head self-attention (MHSA) in MDM cooperate to generate the representation for the WSI. In this process, the \textit{IQ} and variational query \textit{VQ} successfully build up the connections between instances and significantly improve the model's robustness toward heterogeneous tumors. At last, to further enforce constraints in the feature space and stabilize the training process, we adopt a memory-based contrastive loss, which is practicable for WSI classification with a single input of each iteration. 
We conduct experiments on three computational pathology datasets, e.g., CAMELYON16, TCGA-NSCLC, and TCGA-RCC datasets. The superior accuracy and AUC demonstrate the excellence of our proposed MDMIL over other state-of-the-art methods. 
\end{abstract}


\section{Introduction}
\label{sec:introduction}
Whole slide imaging, which refers to scanning and converting a complete microscope slide to a digital Whole Slide Image (WSI), is an efficient technique for visualizing tissue sections in disease diagnosis, medical education, and pathological research \cite{cornish2012whole,pantanowitz2011review}. 
In recent years, with the development of artificial intelligence and big data technology, computational pathology based on WSIs has attracted more research attention than ever due to their clinical and business value in precision medicine. Huge successes have been made in combing medical image analysis, machine learning, and deep learning \cite{lu2021data,bulten2020automated,campanella2019clinical,zeng2021deep,yu2016predicting,wang2021heal,wang2022cell}.

Typically, the gigapixel WSIs with a size of about 40,000 $\times$ 40,000 are computationally infeasible with current hardware. Hans \textit{et al.} \cite{pinckaers2021detection} train an end-to-end deep neural network taking the entire WSIs as input directly by utilizing a streaming implementation of convolutional layers. Even with the specific designs, this method still suffers from relatively low accuracy and high computational requirements. 
Therefore, patch-based processing approaches \cite{cruz2014automatic,hou2016patch,maksoud2020sos,mousavi2015automated}, which divide each WSI into thousands of small patches and utilize the neural networks for examination, have become a mainstream practice for high-dimensional pathology images. 
Considering the patch-level labelling of WSIs for pathologists is time-consuming and challenging, weakly supervised multiple instance learning (MIL), which eliminates the laborious and expensive process of instance labelling by assuming an entire bag with one label, is dominant in this area \cite{ilse2018attention,shao2021transmil,li2021dual,campanella2019clinical,kanavati2020weakly}.

Currently, most MIL methods are designed under the independent and identical distribution (\textit{i.i.d.}) hypothesis, which ignores either the correlations between instances or the heterogeneity of tumors. For example, Ilse \textit{et al.} \cite{ilse2018attention} and Li \textit{et al.} \cite{li2021dual} propose the DeepMIL and DSMIL, which introduce the attention and cross-attention mechanisms into the modelling. 
However, they fail to build connections between instances, and the aggregation weights generated with only internal information are not reliable enough. 
Shao \textit{et al.} \cite{shao2021transmil} proposes TransMIL, which is the first to consider the instance correlations in WSI classification. They utilize a single classification token together with the multi-head self-attention (MHSA) to collect valuable features from the whole bag. Considering the heterogeneity of tumors, the trained model has a risk of overfitting the training data and fails to generalize well on diversified testing data.

To tackle the above issues, we propose multiplex-detection-based multiple instance learning (MDMIL) for WSI classification. MDMIL is constructed by the internal query generation module (IQGM) and multiplex detection module (MDM) and assisted by the memory-based contrastive loss during the training phase. 
Specifically, IQGM gives probabilities of instances on deep-transferred instance features through a linear classification layer and generates the internal query \textit{IQ} by aggregating highly reliable features after the distribution analysis. 
Then, MDM, which consists of the multiplex-detection cross-attention (MDCA, cross-attention module) and multi-head self-attention \cite{vaswani2017attention} (MHSA, self-attention module), aims to generate the final representations for WSIs. 
Here, MDCA establishes the connections between instances and improves the model's generalization ability towards heterogeneous tumors through multiplex-detection strategy (jointly internal and variational queries). 
Then, MHSA establishes the communications of the representations corresponding to different subtypes, reducing overlapping in the feature space and highlighting the critical characteristics. In addition, inspired by the recent success of self-supervised learning work MoCo \cite{he2020momentum}, we ingeniously adopt memory-based contrastive loss into the training phase. 
It greatly relieves the instability training issue with a single WSI input in each iteration and connects different WSIs in the feature space. To the best of our knowledge, our paper is the first to adopt memory-based metric learning to improve the WSI classification performance.

We evaluate our proposed MDMIL for weakly supervised WSI classification on three benchmarks, including Camelyon16, TCGA-RCC, and TCGA-NSCLC. The experiment results demonstrate the state-of-the-art performance of MDMIL over other methods by a large margin.

\begin{figure*}[t]
  \centering
   \includegraphics[width=1.0\linewidth]{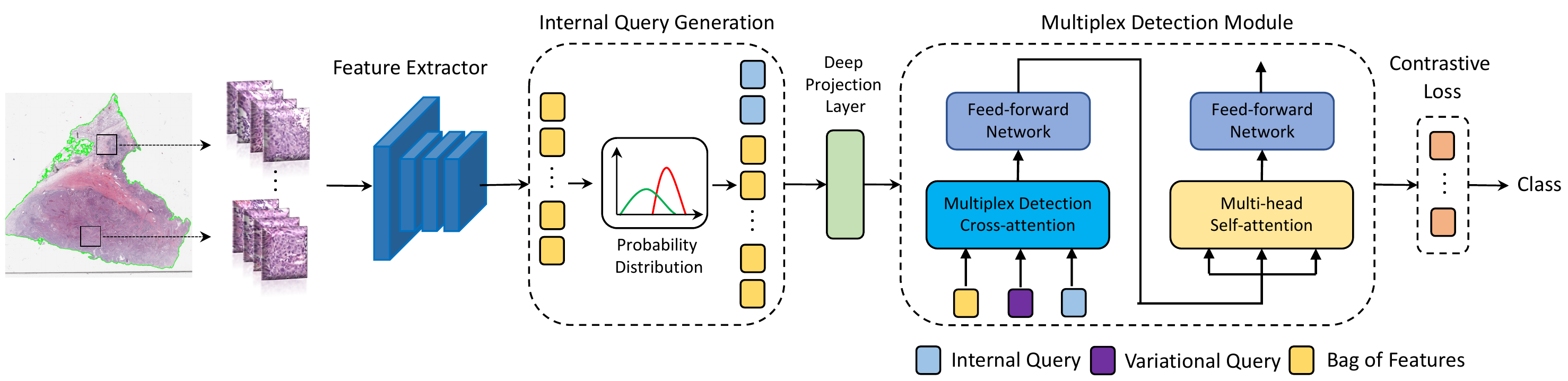}
   \caption{Overview of the proposed multiplex-detection-based multiple instance learning network. We first adopt a segmentation algorithm to discard the background and crop the tissue regions into patches. Secondly, patches are embedded into feature vectors with a pre-trained feature extractor. Then, the bag of features is fed into the internal query generation module (IQGM), deep projection layer (DPL), and multiplex detection module (MDM) to generate the final representation of the WSI. Finally, one linear classification layer outputs the estimated probability of WSI. 
   }
   \label{Fig1}
\end{figure*}

\section{Related Work}
\subsection{Application of MIL in WSI Classification}
There are roughly two categories of MIL in WSI classification: instance-level algorithms and embedding-level algorithms. The instance-level algorithms \cite{kanavati2020weakly,campanella2019clinical,xu2019camel,lerousseau2020weakly} assign the WSI label to the cropped patches directly during the training process. Then, they select the top instances for aggregation and prediction. Apparently, this strategy will introduce a large number of noise labels and result in the trained model with bad convergence conditions and performance. Besides, since only a small number of patches of each slide can join the training, thus a large number of WSIs are required. 
As for the embedding-level algorithms, each patch in the entire WSI is mapped to fixed-length feature vectors; then, the feature vectors are aggregated through aggregation operation (e.g., max-pooling, average pooling, designed aggregation module). 
Many attention-based MIL algorithms \cite{ilse2018attention,tomita2019attention,naik2020deep,lu2021data} have recently been proposed due to the prevalence of attention mechanisms in natural image and natural language processing. Specifically, they calculate an attention score for each instance with trainable parameters and then generate the final representation with a weighted sum. These methods eliminate the requirement of patch-level labels and predict with global characteristics rather than selected patches. Besides, feature clustering methods \cite{sharma2021cluster,wang2019weakly,xie2020beyond} get the final representations by calculating and aggregating the cluster centroids of all the feature embeddings. Recently, transformer architecture \cite{vaswani2017attention}, which has the context-aware ability, is also adopted in MIL to learn the correlation between instances. Shao \emph{et al.} \cite{shao2021transmil} also propose a Transformer-based WSI classification model, which comprehensively considers the correlations among different instances within the same bag. However, that model relies on the trainable parameters with less prior knowledge from the instances directly, thus failing to achieve remarkable performance progress. 

\subsection{Attention in Deep Learning}
In neural networks, the attention mechanism is a technique that mimics human cognition. Initially, the attention mechanism was used to extract meaningful information from sentences in machine translation \cite{bahdanau2014neural}; the attention weights intuitively show how the network adjusts its focus according to context. With its success in natural language processing, many researchers employ it to tackle natural image tasks, \textit{e.g.}, image classification \cite{hu2018squeeze}, object detection \cite{wang2018non,zhu2018attention,zhang2018progressive}, person re-identification \cite{wang2021robust,wang2020robust} and object segmentation \cite{wang2018non,fu2019dual}. 
Essentially, these methods give appropriate weights to the spatial, channel or temporal dimensions for emphasizing valuable features and ignoring the inferior ones. 
Recently, many MIL algorithms have adopted attention mechanisms for feature aggregation and achieved good performance. 

As one of the most well-known architectures of attention mechanisms, Transformer prevails in many different areas \cite{vaswani2017attention,devlin2018bert,dai2019transformer,wang2022feature,yuan2021tokens} due to its impressing ability for long-range feature extraction. 
The original Transformer \cite{vaswani2017attention} in NLP is a novel architecture aiming to solve sequence-to-sequence tasks while handling long-range dependencies with ease. It relies entirely on self-attention to compute its input and output representations without using sequence-aligned RNNs or convolution. 
The Vision Transformer (ViT) \cite{yuan2021tokens} emerged as a competitive alternative to convolutional neural networks (CNNs) that are currently state-of-the-art in computer vision and, therefore widely used in different image recognition tasks. BotNet \cite{srinivas2021bottleneck}, a conceptually simple yet powerful backbone architecture, incorporates self-attention and CNNs together and achieves superior performance for multiple computer vision tasks, including image classification, object detection and instance segmentation. In this paper, our proposed MDMIL is also derived from the Transformer architecture.

\section{Methodology}
\label{Methodology}
This section briefly introduces the proposed multiplex-detection-based multiple instance learning (MDMIL). The overall architecture is illustrated in Fig.\ref{Fig1}. 
Firstly, we segment the tissue region of each slide through an automated segmentation algorithm and divide it into patches with a fixed size. Secondly, we adopt the Imagenet pre-trained model to extract features of patches. Following \cite{lu2021data,shao2021transmil}, we only adopt the first \textit{Convolution Block} and the first three \textit{Residual Blocks} of the ResNet50 \cite{he2016deep} as the extractor; therefore each patch is embedded in a 1024-dimensional feature vector, and the bag of instances can be represented as $F \in \mathbb{R}^{n\times1024}$. After converting all tissue patches into low-dimensional feature embeddings, training and inference can occur in the low-dimensional feature space instead of the high-dimensional pixel space. Thirdly, the proposed IQGM analyzes the probability distribution of instances and generates the \textit{IQ} for MDM. Next, the MDM detects the critical instances and aggregates the final representations for the WSI. Meanwhile, we utilize the memory-based contrastive loss in the training phase to enforce distance constraints in the feature space. Details of each module will be presented in the following section.

\subsection{MIL Formulation}
MIL takes a group of training samples as a bag with multiple instances. The bag labelled with positive contains at least one positive instance, whereas the negative bag must only contain negative ones. Since the instance-level labels are unknown, we refer to MIL as weakly supervised learning. In the case of binary classification task, we have $B=\{(x_1, y_1),...,(x_n, y_n)\}$, where $y_i \in \{0, 1\}$, and $x_i \in \mathbb{R}$. The bag label of $B$ is formulated as:
\begin{equation}
    L(B)=\left\{\begin{matrix}
        0
        &iff \sum_{i=1}^{n}y_i = 0\\
        1
        &otherwise
\end{matrix}\right.
\end{equation}
Typically, MIL consists of three steps: (1) transforming instances to feature embeddings using a function $f$, (2) aggregating transformed instances with symmetric (permutation-invariant) function $\sigma$, (3) and projecting the aggregated features with the function $g$. The process can be formulated as:
\begin{equation}
    MIL(B) = g(\sigma(f(x_1),...,f(x_n))).
\end{equation}

\begin{algorithm}[htb] 
	\caption{Internal Query Generation Module} 
	\label{alg:Framwork} 
	\begin{algorithmic}[1] 
		\renewcommand{\algorithmicrequire}{\textbf{Input:}}
		\renewcommand{\algorithmicensure}{\textbf{Output:}}
		\REQUIRE ~~\\
		A bag of feature embeddings $F={f_{1},...,f_{n}}$, where $F \in \mathbb{R}^{n \times d}$ and $f_{i} \in \mathbb{R}^{1 \times d}$.
		\ENSURE ~~\\
		Reliable \textit{IQ} of each subtype;
		
		\STATE Get the probability distribution of instances through a classification layer; \\
		$P \longleftarrow softmax (cls_1(\{f_{1},...,f_{n}\}))$;
		
		\STATE Calculate the confidence factor for each subtype through top $K_1$ instances; \\
		$CF \longleftarrow \{cf_i=\bar{m_i}-\sigma_i; i=1,...,N \}$;
		
		\STATE Estimate the internal query;\\
		if $cf_{max} - \beta  > \forall\{cf_1,...,cf_N;index \ne max \}$:\\
		\quad Average top $K_1$ and $K_2$ instances for confident and other subtypes, respectively \\
		\quad $IQ=\{q_{rb}, q_i; index \ne max\}$\\
		else:\\
		\quad Average top $K_2$ instances for all the subtypes \\
		\quad $IQ=\{q_1,...,q_N\}$
	\end{algorithmic}
	\label{algorithm}
\end{algorithm}

\subsection{Internal Query Generation Module}
Our proposed IQGM aims to generate the internal query \textit{IQ} as the internal detector for the subsequent MDM. 
In DSMIL \cite{li2021dual}, they utilize a classification layer $cls$ together with a $softmax$ function and a $max-pooling$ operation to retrieve features with the highest probability corresponding to each subtype. 
However, this strategy has two obvious shortcomings: (1) the prediction accuracy of $cls$ is relatively low, and the absolute predictive power of the whole model will be logically limited at the upper bound of the $cls$'s expression; (2) due to the application of deep transfer learning and the heterogeneity of tumors, the retrieved features will be short of class-level representative and have much patch-specific information. 

A straightforward approach to tackle the above issues is averaging the top instance features of each subtype. However, this approach will introduce false-negative information with a significant chance, affecting the convergence of the model and the stability of the model training.  

Following DSMIL\cite{li2021dual}, we first adopt a classification layer $cls$ and a $softmax$ function to get the probability distribution of instances as:
\begin{equation}
    P = softmax(cls_1(\{f_{1},...,f_{n}\})).
\end{equation}
Then, we define a confidence factor $cf_i$ for subtype $i$ through the mean value $\bar{m}_i$ and standard deviation $\sigma_i$ of the top $K_1$ instances:
\begin{equation} 
    cf_i=\bar{m_i}-\sigma_i;
    \bar{m_i}=\frac{1}{K_1}\sum_{j=1}^{K_1}p_{ij};\\ \sigma_i=\sqrt{\frac{\sum_{j=1}^{N}(p_{ij}-\bar{m}_i)^2}{K_1}},\\
\end{equation}
Suppose the maximum one $cf_{max}$ satisfies the requirement:
\begin{equation}
    cf_{max} - \beta >\forall\{cf_1,...,cf_N;index \ne max \},
\end{equation}
where $\beta \in [-1,1]$ is a hyperparameter for alleviating the instance imbalance issue. In that case, we define the subtype of $cf_{max}$ as a confident one and aggregate the top $K_1$ instance features with mean function as reliable internal query $q_{rb}$. As for other subtypes, we only aggregate the top $K_2$ instance features as $q$. If no $cf_{max}$ exits, we suspect the reliability of the output probability and only average the top $K_2$ instance features. Here, $K_1$ and $K_2$ are not fixed numbers; they are the multiplication of the bag length and predefined ratio hyperparameter $r_1, r_2 \in [0,1]$. Typically, $r_1$ is bigger than $r_2$ to involve more instances in the analysis. We will discuss parameters along with the datasets in Sec.\ref{experiments}. The pipeline of the algorithm is shown in Alg.\ref{algorithm}.

Through the proposed IQGM, we can generate a reliable \textit{IQ} with less false-positive information, thus laying a good foundation for the following prediction.

\begin{figure*}[t]
  \centering
   \includegraphics[width=0.9\linewidth]{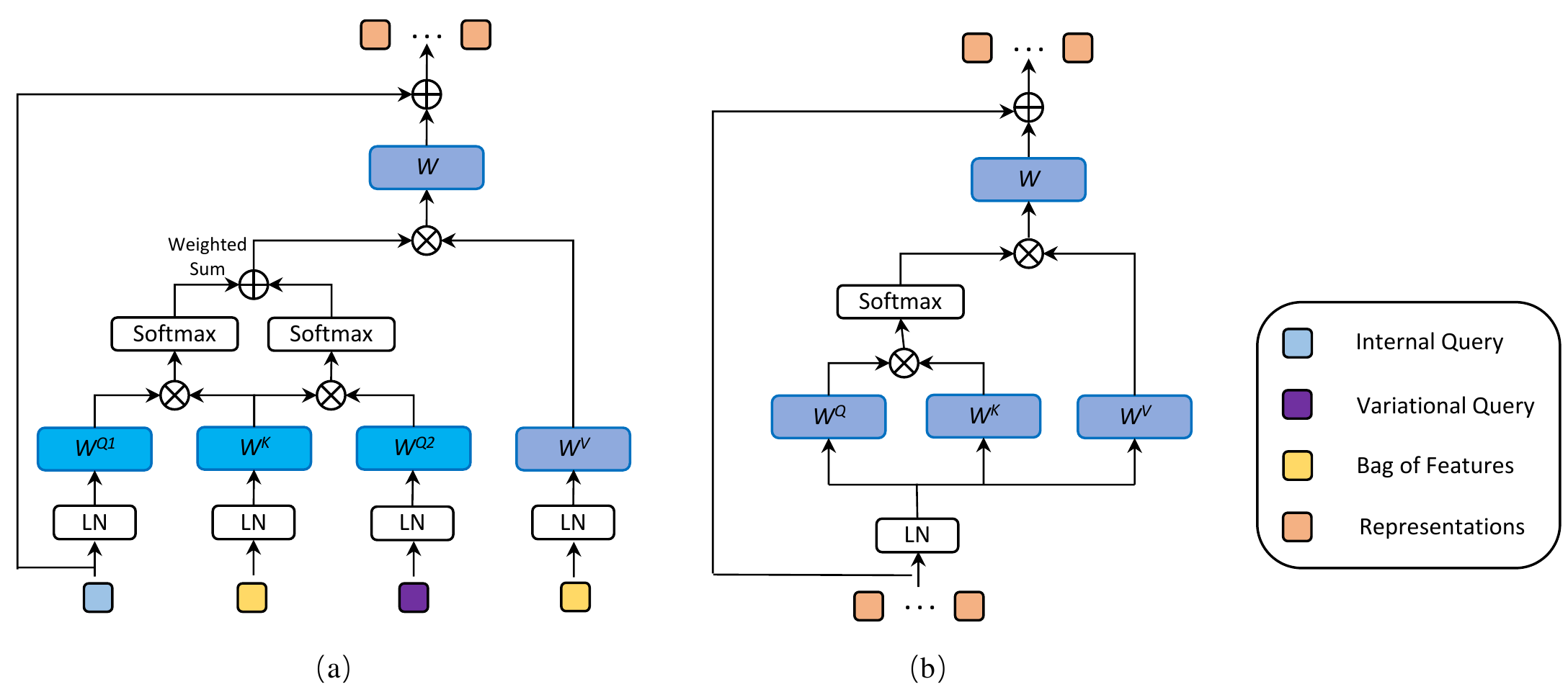}
   \caption{Overview of the proposed multiplex-detection cross attention (MDCA) (a) and traditional multi-head self-attention (MHSA) (b). The main differences between the two modules lay in the input and the attention strategy. There are two \textit{Query} in MDCA, and MHSA arises \textit{Query}, \textit{Key} and \textit{Value} from one input. 
   }
   \label{Fig2}
\end{figure*}

\begin{figure}[t]
  \centering
   \includegraphics[width=0.6\linewidth]{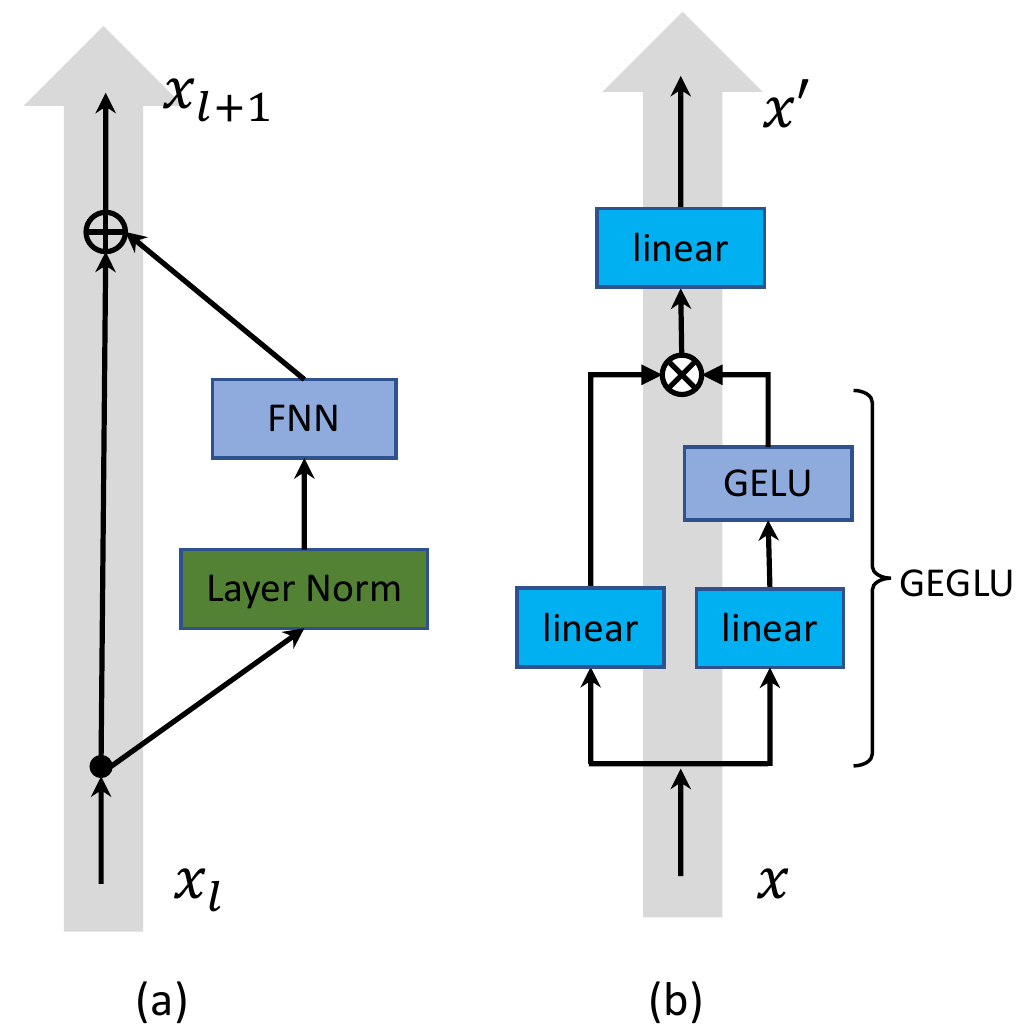}
   \caption{Overview structure of the feed-forward network (FFN) (a) and its FFN module (b). }
   \label{Fig3}
\end{figure}

\subsection{Multiplex Detection Module}
The proposed MDM aims to detect the critical instances that trigger the prediction and generate the final representations of WSIs. Previous methods adopt either an unreliable internal query \cite{li2021dual} or a single variational query \cite{shao2021transmil} for detection, failing to create a robust algorithm due to their inferior feature mining ability. To tackle these issues, we integrate the two strategies within one module, enabling the proposed module with superior performance. As shown in Fig \ref{Fig1}, MDM consists of one multiplex-detection cross-attention (MDCA), one multi-head self-attention (MHSA), and two feed-forward networks (FFN)\cite{xiong2020layer}. We illustrate the MDCA, MHSA and FFN in Fig.\ref{Fig2} (a), Fig.\ref{Fig2} (b) and Fig.\ref{Fig3}, respectively.

Before the MDM, we employ a deep projection layer (DPL) constructed by one nonlinear projection unit and one linear projection layer to achieve feature recalibration. The nonlinear projection unit consists of one fully connected layer, one layer normalization ($LN$) \cite{ba2016layer}, and one $ReLU$ activation function. DPL is essential in the network because it can project features of similar tissue sections to a relatively compact cluster in the feature space, facilitating feature mining in the subsequent attention operations.

Essentially, MDCA is a modified cross-attention module based on the standard architecture of the MHSA. MDCA has three inputs: internal query \textit{IQ} from IQGM, trainable variational query \textit{VQ}, and the bag of features $F'$ after the DPL. Given the feature vector, \emph{query one} $Q_1$ arises from the \textit{IQ}, \emph{query two} $Q_2$ arises from the \textit{VQ}, and \emph{key} $K$ and \emph{value} $V$ arise from the $F'$. Formally, 
\begin{equation}
\begin{aligned}
    Q_1 = LN(IQ) W^1, Q_2 = LN(VQ) W^2, \\
    K = LN(F') W^3, V = LN(F') W^4,
\end{aligned}
\end{equation}
where $W \in \mathbb{R}^{d \times d'}$ is linear projection. We retain the multi-head strategy in MHSA and split features into $m$ parts along the channel dimension, and get $Q^1, Q^2 \in \mathbb{R}^{N \times m \times (d/m)}$ and $K, V \in \mathbb{R}^{n \times m \times (d/m)}$. Next, we calculate two attention matrixes ($mt^1$ and $mt^2$) based on $Q^1$ and $Q^2$, respectively. Formally, the function for each matrix can be presented as:
\begin{equation}
   mt_{i,j}=\frac{exp(\beta_{i,j})}{\sum_{j=1}^{n}exp(\beta_{i,j})}, \quad \beta_{i,j}=\frac{Q_iK_j}{\sqrt{d_k}},
\end{equation}
where $\sqrt{d_k}$ is a scaling factor and, $i$ and $j$ present the indexes of $query$ and $key$, respectively. Each element of the attention matrix indirectly indicates the similarity between \textit{query} and \textit{key}. 
We take the weighted sum of $mt_1$ and $mt_2$ as the final attention matrix $m'$. 
We define the weights of the two matrixes according to the reliability of \textit{IQ}. If the critical instances cover a large area of the WSI and can be found easily, \textit{IQ} will be assigned more attention for fast model convergence. Otherwise, we will assign more attention to the \textit{VQ}. The aggregation process can be defined as:
\begin{equation}
\begin{aligned}
    f_{ag} = MDCA(Q^1, Q^2, K, V) \\
    = (\alpha \cdot mt^1 + (1-\alpha) \cdot mt^2) \times V.
\end{aligned}
\end{equation}
With the cross-attention between \textit{IQ} and $K$, we successfully establish the connections between each instance and a collection of potential critical ones. Through the cross-attention between \textit{VQ} and $K$, we further supplement other features to improve the model's robustness toward heterogeneous tumors. 
Then, we conduct a projection operation with a linear layer. At last, we perform the residual operation between the attention output and \textit{IQ} to avoid the gradient vanishing. The final output of MDMC is $f_{ag} \in \mathbb{R}^{N \times d}$, where $N$ is the number of subtypes. 

After the cross-attention operation, we employ the post-layer normalization feed-forward network (FFN) \cite{xiong2020layer} to conduct a non-linear transformation. As shown in Fig.\ref{Fig3} (a), a residual connection is applied on the original and transformed features. As for the FFN module Fig.\ref{Fig3} (b), we first adopt a linear layer to expand the feature dimension, and then feed the expanded features into GEGLU \cite{shazeer2020glu}. GEGLU projects the features into two parts and adopts the activation function \textit{GELU} \cite{hendrycks2016gaussian} to nonlinear the features. Then we adopt the multiplication of the two features as output. Functionally, GEGLU can be expressed by:
\begin{equation}
    GEGLU(f_{ag}, W, V, b, c) = GELU(f_{ag}W + b) \cdot (f_{ag}V + c),
\end{equation}
where $W,V$ are the transform matrixes, and $b,c$ are the biases. A linear layer is applied on the features for recalibration.

Next, we employ the MHSA \cite{vaswani2017attention} (as shown in Fig.\ref{Fig2} (b)) to establish the communications of the representations corresponding to different subtypes. The self-attention module does not require any external information and can compute independently.
Through the self-attention operation, we can further disentangle representations and guarantee less overlapping in the feature space. Finally, we utilize another nonlinear transformation FFN to recalibrate the representations. 
After getting the features that correspond to all subtypes, we simply average all the features to get the final representation of the WSI.

We widely adopt $LN$ instead of batch normalization \cite{ioffe2015batch} in this module because $LN$ directly estimates the normalization statistics of the input samples within a hidden layer so that the normalization does not introduce any new dependencies between the training cases. The model can optimize well with fewer samples in each mini-batch.

\subsection{Memory-based Contrastive Loss}
Metric learning is very effective for optimizing deep neural networks. However, since WSIs have different numbers of instances and only one bag of features can be fed into the neural network in each iteration; thus metric learning by referring to other samples within the mini-batch is infeasible. 
Recently, memory-based methodologies have been widely adopted in self-supervised learning \cite{he2020momentum} and other computer vision tasks \cite{wang2022feature}. Inspired by these methods, we adopt a memory-based contrastive loss to promote network performance, which can implement constraints in the feature space by referring to the global information.

There are two steps for the loss function: memory initialization at the beginning of the algorithm and memory update throughout the training process. The memory is initialized with the feature centers in the training set. Firstly, we perform the forward computation to get the labelled subtype representation. Then, we conduct average and l2 normalization operations for features of each subtype separately to get the feature centers $C=\{c_1,...,c_N\}$. As for the memory update, the $k$-th center $c_k$ is updated during training by a momentum controlled as:
\begin{equation}
    c_k = m c_k + (1 - m)f_i,
\end{equation},
where $m$ is the momentum coefficient and $f_i$ is the labelled subtype representation.
Functionally, the \emph{Contrative Loss} is:
\begin{equation}
    \mathcal{L}_{CL}=-log\frac{exp(<f,c_i>/\tau )}{\sum_{j}^{N}exp(<f,c_j>/\tau)},
\end{equation}
where $\tau$ is a predefined temperature parameter and $c_i$ represents the feature center with an identical subtype.

\subsection{Overall Loss}
There are three loss functions during the training phase: two cross-entropy loss ($\mathcal{L}_{CEL}$) and one contrastive loss ($\mathcal{L}_{CL}$). We calculate the $\mathcal{L}_{CEL}$ with the classification output $P_1$ in IQGM and the final classification layer $P_2$. 
Notably, we adopt a $max-pooling$ operation on $P_1$ to get the instance with the highest score for loss calculation. 
Functionally, $\mathcal{L}_{CEL}$ can be presented as:
\begin{equation}
    \mathcal{L}_{CEL}=-y_ilog(\frac{exp(W_if_i)}{\sum_{j=1}^{N}exp(W_jf_j)}),
\end{equation}
where $W$ is a linear projection matrix, $y_i$ is the corresponding label and $N$ is the total number of classes. 

Therefore, the overall loss functions can be expressed as: 
\begin{equation}
\begin{aligned}
    \mathcal{L}_{Final}=\mathcal{L}_{CEL1}(max-pooling(P_1), y) \\
    + \mathcal{L}_{CEL2}(P_2, y) + \alpha\mathcal{L}_{CL},
\end{aligned}
\end{equation}
where $\alpha$ is a predefined soft parameter and is 0.5. 

\begin{table*}
  \centering
  \normalsize
  \begin{tabular}{l*{6}{c}}
    \toprule
    & \multicolumn{2}{c}{CAMELYON16} & \multicolumn{2}{c}{TCGA-NSCLC} & \multicolumn{2}{c}{TCGA-RCC} \\
    \cmidrule(lr){2-3} \cmidrule(lr){4-5} \cmidrule(lr){6-7}
    & Accuracy & AUC & Accuracy & AUC & Accuracy & AUC \\
    \midrule
    Mean-pooling & 0.6389 & 0.4647 & 0.7282 & 0.8401 & 0.9054 & 0.9786 \\
    Max-pooling & 0.8062 & 0.8569 & 0.8593 & 0.9463 & 0.9378 & 0.9879 \\
    ABMIL \cite{ilse2018attention} & 0.8682 & 0.8760 & 0.7719 & 0.8656 & 0.8934 & 0.9702\\
    PT-MTA \cite{li2019patch} & 0.8217 & 0.8454 & 0.7379 & 0.8299 & 0.9059 & 0.9700\\
    MIL-RNN \cite{campanella2019clinical} & 0.8450 & 0.8880 & 0.8619 & 0.9107 & - & -\\
    DSMIL \cite{li2021dual} & 0.7985 & 0.8179 & 0.8058 & 0.8925 & 0.9294 & 0.9841\\
    CLAM-SB \cite{lu2021data} & 0.8760 & 0.8809 & 0.8180 & 0.8818 & 0.8816 & 0.9723\\
    CLAM-MB \cite{lu2021data} & 0.8372 & 0.8679 & 0.8422 & 0.9377 & 0.8966 & 0.9799\\
    TransMIL \cite{shao2021transmil} & 0.8837 & 0.9309 & 0.8835 & \textcolor{red}{0.9603} & 0.9466 & 0.9882\\
    \hline
    Ours & \textcolor{red}{0.9158} & \textcolor{red}{0.9669} & \textcolor{red}{0.9052} & 0.9596 & \textcolor{red}{0.9693} & \textcolor{red}{0.9944}\\
    \bottomrule
  \end{tabular}
  \caption{Comparison with other state-of-the-art methods on CAMELYON16, TCGA-NSCLC and TCGA-RCC datasets. We take accuracy and AUC for evaluation.}
  \label{table1}
\end{table*}

\begin{table}
  \centering
  \setlength\tabcolsep{3pt}
  \footnotesize
  \begin{tabular}{ccccccccc}
    \toprule
    \multicolumn{5}{c}{} & \multicolumn{2}{c}{TCGA-NSCLC} & \multicolumn{2}{c}{TCGA-RCC}\\
    \midrule
    & MDM & DPL & IQGM & $\mathcal{L}_{CL}$ & Accuracy & AUC & Accuracy & AUC \\
    0 & \XSolidBrush & \XSolidBrush & \XSolidBrush & \XSolidBrush & 0.8038 & 0.8918 & 0.7609 & 0.9108 \\ 
    1 & \Checkmark & \XSolidBrush & \XSolidBrush & \XSolidBrush & 0.8341 & 0.9200 & 0.8930 & 0.9884 \\ 
    2 & \Checkmark  & \Checkmark  & \XSolidBrush & \XSolidBrush & 0.8768 & 0.9432 & 0.9412 & 0.9906 \\ 

    3 & \Checkmark & \Checkmark  & \Checkmark & \XSolidBrush & 0.8957 & 0.9442 & 0.9305 & 0.9913 \\
    
    4 & \Checkmark & \Checkmark  & \XSolidBrush & \Checkmark & 0.8815 & 0.9365 & 0.9519 & 0.9950 \\
    
    5 & \Checkmark & \Checkmark  & \Checkmark & \Checkmark & 0.8957 & 0.9515 & 0.9679 & 0.9948 \\ 
    \bottomrule
  \end{tabular}
  \caption{Ablation study on TCGA-NSCLC and TCGA-RCC datasets. Specifically, we evaluate the proposed MDM, IQGM, and CL. We take accuracy and AUC for evaluation.}
  \label{table2}
\end{table}

\begin{table}
  \centering
  \setlength\tabcolsep{3pt}
  \begin{tabular}{ccccccc}
    \toprule
    \multicolumn{1}{c}{} & \multicolumn{2}{c}{CAMEYLON16} & \multicolumn{2}{c}{TCGA-NSCLC} & \multicolumn{2}{c}{TCGA-RCC}\\
    \midrule
    $\beta$ & Accuracy & AUC & Accuracy & AUC & Accuracy & AUC \\
    -0.1 & \textcolor{red}{0.9158} & \textcolor{red}{0.9669}  & 0.8830 & 0.9374 & 0.9519 & \textcolor{red}{0.9951} \\ 
    -0.05 & 0.9141 & 0.9664  & 0.8912 & 0.9429 & 0.9626 & 0.9950 \\ 
    0.0 & 0.8750 & 0.9357  & 0.8957 & 0.9515 & 0.9679 & 0.9948 \\ 
    0.05 & 0.8672 & 0.9445 & \textcolor{red}{0.9052} & \textcolor{red}{0.9596} &  \textcolor{red}{0.9693} & 0.9944  \\ 
    0.1 & 0.8359 & 0.9057 & 0.8830 & 0.9374 & \textcolor{red}{0.9693} & 0.9944  \\
    \bottomrule
  \end{tabular}
  \caption{Analysis of the $\beta$ in MDM on CAMELYON16, TCGA-NSCLC and TCGA-RCC datasets. We take accuracy and AUC for evaluation.}
  \label{table3}
\end{table}

\section{Experiments and Results}
\label{experiments}
In this paper, we demonstrate the superior performance of the proposed MDMIL over three public datasets: The Cancer Genome Atlas non-small cell lung cancer (TCGA-NSCLC), TCGA renal cell carcinoma (TCGA-RCC) and CAMELYON16.

\subsection{Dataset}
TCGA-NSCLC dataset has two subtypes: lung squamous cell carcinoma (TCGA-LUSC) and lung adenocarcinoma (TCGA-LUAD). There are 993 diagnostic WSIs, including 507 LUAD slides from 444 cases and 486 LUSC slides from 452 cases. After processing, the mean number of patches extracted per slide at $\times$20 magnification is 11,038.

The TCGA-RCC dataset has three subtypes: kidney chromophobe renal cell carcinoma (TCGA-KICH), kidney renal clear cell carcinoma (TCGA-KIRC) and kidney renal papillary cell carcinoma (TCGA-KIRP). There are 884 diagnostic WSIs, including 111 KICH slides from 99 cases, 489 KIRC slides from 483 cases, and 284 KIRP slides from 264 cases. After processing, the mean number of patches extracted per slide at $\times$20 magnification is 12,386.

CAMELYON16 is a public dataset about metastasis in breast cancer. There are 270 WSIs in the training set, including 159 normal tissues and 111 tumor tissues. As for the testing set, there are in total of 130 WSIs. After processing, the mean number of patches extracted per slide at $\times$20 magnification is 9,752.

\subsection{Implementation Details and Evaluation Metrics}
Here, we train our network in an end-to-end fashion through the Adam optimizer with a weight decay of 5e-3. The learning rate is initialized as 2e-4 with cosine learning rate decay. In each iteration, only one bag of features from one WSI will be fed as input data. To increase the diversity of input data, we randomly discard 0 $\sim$ 10\% instances as an augmentation strategy in the training phase. For the two TCGA datasets, we only train the model for 100 epochs. As for the CAMELYON16 dataset, considering the fewer critical instances in each WSI, we train the model for 200 epochs. 

In this paper, we report the area under curve (AUC) scores and accuracy for evaluation. AUC is a graph showing the performance of a classification model at all classification thresholds. Accuracy gives the percentage of correct classifications. 

\subsection{Experiment Setup}
Following \cite{shao2021transmil}, we crop each WSI into a series of 256 $\times$ 256 non-overlapping patches and discard the background patches (saturation $<$ 15). For TCGA datasets, we randomly split the whole dataset into non-overlapping training, validation, and testing datasets with a ratio of 60\%, 15\%, and 25\%. Since CAMELYON16 dataset has an official testing set, we take 90\% and 10\% WSIs of the training set for training and validation, respectively. 
The best-performing model on the validation set will be taken for testing. 

\begin{figure*}[t]
  \centering
   \includegraphics[width=1.0\linewidth]{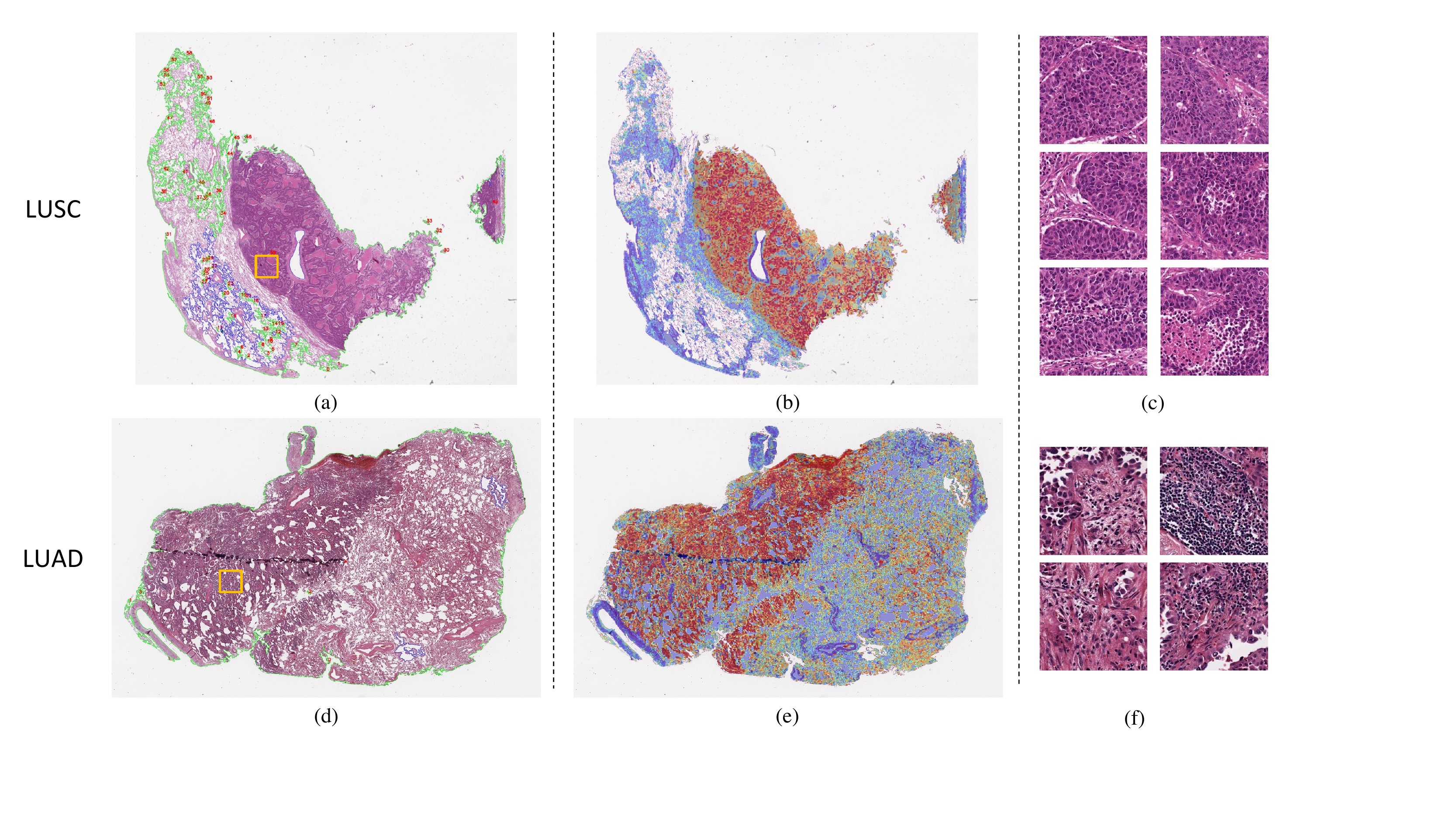}
   \caption{Attention visualization of both LUSC and LUAD WSIs. The first (a,d) and second (b,e) columns present the original WSIs segmented with green and blue curves, and images mapped with generated attention scores from MDCA, respectively. The last column (c,f) gives high score instances cropped by the yellow boxes on original WSIs.  
   }
   \label{Fig4}
\end{figure*}

\begin{figure}[t]
  \centering
   \includegraphics[width=1.0\linewidth]{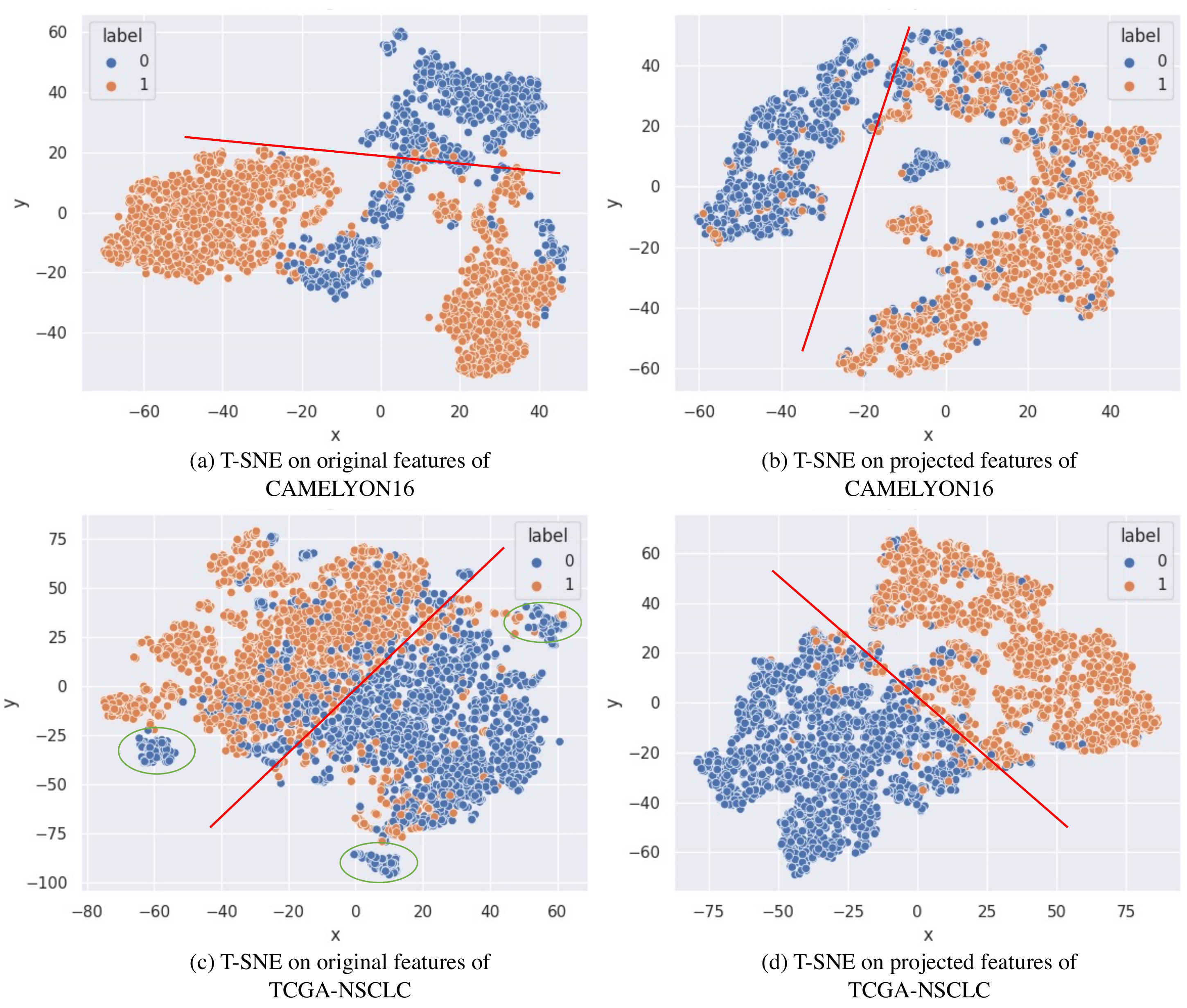}
   \caption{Feature distribution before and after DPL on CAMELYON16 (a,b) and TCGA-NSCLC (c,d) datasets. Here, we take T-SNE \cite{van2008visualizing} for visualization in the feature space. It is obvious that DPL helps to separate features of different subtypes apart, alleviating the learning pressure of subsequent modules. 
   }
   \label{Fig5}
\end{figure}

\subsection{Results on WSI classification}
We evaluate our proposed model on both detection and subtype classification datasets. For the detection dataset, e.g., CAMELYON16, the positive WSI contains metastases while the negative ones don't contain metastases. As for subtype classification datasets, e.g., TCGA-NSCLC and TCGA-RCC, each subtype has its unique pattern. We present all the results in Table.\ref{table1}.

\textbf{CAMELYON16:}
Since the positive slides only contain a small portion of metastasis tissue (roughly $<$ 10\% of the tissue area), CAMELYON16 is one of the most challenging datasets to testify to the effectiveness of the MIL algorithm. Only algorithms capable of feature mining can detect critical instances and make accurate predictions. Considering the instance distribution of the dataset, we set $r_1$ and $r_2$ as 0.05 and 0.01, respectively. 
As shown in Table.\ref{table1}, our method achieve 0.9158 and 0.9669 on accuracy and AUC. Especially on AUC, we surpass TransMIL \cite{shao2021transmil} and CLAM-SB \cite{lu2021data} by 3.6\% and 8.6\%, respectively. TransMIL \cite{shao2021transmil} utilizes the MHSA to bridge the connections between instances and adopt the classification token to aggregate the critical features. The classification token has trouble collecting all the valuable features; thus, the algorithm lacks generalization on the testing set. DSMIL \cite{li2021dual} is based on \textit{i.i.d} hypothesis. They utilize one retrieved internal query to aggregate the bag of features and fail to consider the contextual information and morphologically diversified tumors.

\textbf{TCGA-NSCLC:} 
This dataset has more positive tiles than CAMELYON16, reaching above 80\% per slide. Therefore, the main challenge is distinguishing the fine-grained patterns between subtypes. Here, we set $r_1$ and $r_2$ as 0.1 and 0.01, respectively.
As shown in Table.\ref{table1}, our method surpasses others by at least 2.17\% on accuracy. As for AUC, TransMIL achieves competitive performance, surpassing us only by 0.07\%. We still outperform others by at least 1.4\%. This demonstrates that our MDMIL has a powerful ability for fine-grained feature mining. Even for subtypes with minor differences, MDMIL still can make more accurate decisions than other methods.

\textbf{TCGA-RCC:} 
This dataset, the same as the TCGA-NSCLC, has large areas of tumor region in the positive slide (average area $>$ 80\% per slide). It is a multi-classification problem with three subtypes. 
Apart from the challenge of detecting fine-grained patterns, another one is the data imbalance. KICH only has 70 training slides, much less than 337 slides of KIRC and 193 slides of KIRP. Here, we set $r_1$ and $r_2$ as 0.1 and 0.01, respectively.
As shown in Table.\ref{table1}, our MDMIL can achieve superior performance to others even under such challenges. Especially the AUC, we achieve 99.44\%, quite close to 100\%. This significantly contributes to the memory-based contrastive loss, which consistently enforces distance constraints in the feature space. We also achieve superior accuracy, reaching 96.93\%, much higher than any of the others. Our outstanding performance contributes further steps of the diagnostic algorithm for real-world application.

\subsection{Ablation Study}
\textbf{Analysis of Each Component.}
To verify the effectiveness of each proposed component, we present the ablation study of the internal query generation module (IQGM), deep projection layer (DPL), multiplex detection module (MDM), and memory-based contrastive loss ($\mathcal{L}_{CL}$) on TCGA-NSCLC and TCGA-RCC datasets in Table.\ref{table2}. Here, we set $\beta$ as 0 for a fair comparison. $Model_0$ is the baseline method that utilizes a classification layer together with the $softmax$ and $max-pooling$ operations for the examination. $Model_1$ adds the MDM on top of the transferred features. The same as DSMIL\cite{li2021dual}, the classification and $max-pooling$ operation help to generate the rough $IQ$ for MDM. $Model_2$ adds the DPL before MDM for feature transformation. Comparing $Model_0$, $Model_1$ and $Model_2$, we can see significant improvements in both AUC and Accuracy. Especially on the TCGA-RCC dataset, the Accuracy improves from 76.09\% to 94.12\%, improving by 18\%. From the comparison, we can conclude that our proposed MDM effectively finds critical instances that trigger the bag label and DPL is important for assisting the training. $Model_3$ adds the IQGM in place of the $max-pooling$ operation. It is obvious that with the assistance of IQGM, we can get a more reliable $IQ$ for MDM and indirectly improve the instance mining ability of MDM. $Model_4$ adds $\mathcal{L}_{CL}$ on $Model_2$. With the constraints in the feature space, the Accuracy and AUC improve on both TCGA-NSCLC and TCGA-RCC datasets. $Model_5$ is the MDMIL that consists of IQGM, MDM and $\mathcal{L}_{CL}$. $Model_5$ achieves the best performance, demonstrating that each module can work independently and corporately. 

\textbf{Analysis of $\beta$ in IQGM.}
$\beta$ is a critical hyperparameter in IQGM. By acting as the bias, it can provide prior knowledge for the network. In Table.\ref{table3}, we present the experiment results with different $\beta$, ranging from -0.1 to 0.1. Specifically, we take the CAMELYON16 dataset as an example. As said before, the metastasis region only covers a small portion of WSIs; with the instance imbalance issue, the classification layer trends to give higher probabilities for normal patches and lower probabilities for metastasis ones. 
Therefore, by subtracting $\beta$ ($<0$) to the $cf$ of the metastasis subtype (rather than $cf_{max}$), we can give metastasis patches more attention and avoid critical features being ignored. As for the TCGA dataset, since tumor tissue covers much areas of WSIs, the classification layer in IQGM will not be troubled by the instance imbalance issue within WSIs. $\beta$ is taken as a margin for enhancing the reliability of \textit{IQ}. As shown in Table.\ref{table3}, when $\beta$ is -0.1, we can achieve the best performance on CAMELYON16 datasets. However, when $\beta>0$, the performance drops dramatically, demonstrating that $\beta$ is effective under the instance imbalance circumstances. As for the two TCGA datasets, when $\beta>0$, the performance becomes better and more stable, which proves it can improve the reliability of $IQ$.

\subsection{Interpretability and Whole-slide attention Visualization}
Human readable interpretability of the trained deep neural networks can validate the model's predictive basis aligned with pathologists' knowledge. Besides, attention maps are also a functional tool to assist human-in-the-loop clinical diagnosis. In Fig.\ref{Fig3}, we visualize the attention scores and critical instances of WSIs in TCGA-NSCLC dataset. Here, we take two WSIs of LUSC and LUAD as examples. The first column (a, d) gives the original WSIs after the segmentation algorithm. The green curves circle the tissue region, and the blue ones eliminate the space inside. The second column (b,e) maps the attention scores generated by MDCA onto the images. Specifically, the attention scores are the combination of the attention matrixes ($mt_1$ and $mt_2$) generated by \textit{IQ} and $VQ$. Before mapping, we conduct the normalization operation on the attention matrix to scale the value to 0$\sim$1. In the third column (c,f), we present the high score tissues from the yellow boxes in (a,b). From (b,e), we can see that, although pixel-level or patch-level annotation was never utilized to assist the training of the network, our proposed MDMIL still can highlight the critical instances correlated to the tumor subtype. This finding demonstrates that MDMIL has the potential to be used for meaningful WSI interpretability and visualization in cancer subtyping problems for clinical or research purposes.

In Fig.\ref{Fig4}, we visualize the feature distribution before and after DPL through T-SNE \cite{van2008visualizing} on CAMELYON16 (a,b) and TCGA-NSCLC (c,d) datasets. Specifically, for each WSI, we randomly select 20 instances from the top 20\% ones of the labelled subtype through the first classification layer. 
For the CAMELYON16 dataset Fig.\ref{Fig4} (a), there is a massive difference in feature distribution on positive samples (metastasis) and negative samples (un-metastasis); a simple decision surface can not distinguish them. This finding verifies that the algorithms, e.g., DSMIL, DeepMIL, based on the \textit{i.i.d} without DPL have generalization issues. However, with the application of DPL Fig.\ref{Fig4} (b), although still with independent clusters (circled by green curves), the disorder situation alleviates significantly. As for the TCGA-NSCLC dataset Fig.\ref{Fig4} (c), the features are more distinguishable due to the discrepancy in critical patterns of LUAD and LUSC. However, there remain fewer overlapping and many independent clusters (circled by green curves). Our DPL helps to tackle this issue and separate them in the feature space Fig.\ref{Fig4} (d). 
All in all, DPL is critical in the deep transfer learning architecture and can provide more discriminative features for the following modules.

\section{Summary}
In this paper, we propose the multiplex-detection-based multiple instance learning (MDMIL), which sinuously considers the correlations between instances and the heterogeneity of tumors. Specifically, our proposed MDMIL is constructed by the internal query generation module (IQGM), the multiplex detection module (MDM) and memory-based contrastive loss. The proposed IQGM gives the probability distribution of instances and generates the internal query \textit{IQ}. Then, MDM simultaneously utilizes the pre-computed \textit{IQ} and trainable variational query \textit{VQ} to detect the critical features. With the multiplex constraints, we can quickly build connections between instances and get discriminative representations for WSIs. Next, we adopt a memory-based contrastive loss, practicable for WSI classification with a single input of each iteration. We conduct experiments on three computational pathology datasets, e.g., CAMELYON16, TCGA-NSCLC, and TCGA-RCC datasets. The superior accuracy and AUC demonstrate the superiority of our proposed MDMIL over other state-of-the-art methods.

\bibliographystyle{IEEEtran}  
\bibliography{cite.bib}  
\end{document}